\documentclass{article}

     \usepackage[final]{neurips_2021}

\usepackage[utf8]{inputenc} % allow utf-8 input
\usepackage[T1]{fontenc}    % use 8-bit T1 fonts
\usepackage{hyperref}       % hyperlinks
\usepackage{url}            % simple URL typesetting
\usepackage{booktabs}       % professional-quality tables
\usepackage{amsfonts}       % blackboard math symbols
\usepackage{nicefrac}       % compact symbols for 1/2, etc.
\usepackage{microtype}      % microtypography
\usepackage{xcolor}         % colors

\usepackage[flushleft]{threeparttable}

\title{English2Gbe: A multilingual machine translation model for \{Fon/Ewe\}Gbe}

\author{%
  Gilles Hacheme \\
  Masakhane NLP\\
  Ai4Innov\\
  \texttt{gilles.hacheme@ai4innov.com}
  }

\usepackage{times}
\usepackage{latexsym}

\usepackage[T1]{fontenc}
\usepackage[utf8]{inputenc}

\usepackage{microtype}

\usepackage{multirow}
\usepackage{rotating}

\usepackage{tikz}
\definecolor{mycolor}{RGB}{226,185,252}
\definecolor{mycolor2}{RGB}{140,21,21} 
\definecolor{mycolor3}{RGB}{86,1,141} 
\usepackage{mathtools}
\usetikzlibrary{arrows, decorations.markings,shapes,arrows,fit, positioning}
\usepgflibrary{shapes.arrows}

\usepackage{lscape}

\usepackage[T4, OT1]{fontenc} % loading the fc package for African characters

\begin{document}
\maketitle
\begin{abstract}
    Language is an essential factor of emancipation. Unfortunately, most of the more than 2,000 African languages are low-resourced. The community has recently used machine translation to revive and strengthen several African languages. However, the trained models are often bilingual, resulting in a potentially exponential number of models to train and maintain to cover all possible translation directions. Additionally, bilingual models do not leverage the similarity between some of the languages. 
    Consequently, multilingual neural machine translation (NMT) is gaining considerable interest, especially for low-resourced languages. Nevertheless, its adoption by the community is still limited. This paper introduces English2Gbe, a multilingual NMT model capable of translating from English to Ewe or Fon. Using the BLEU, CHRF, and TER scores computed with the Sacrebleu \citep{post2018call} package for reproducibility, we show that English2Gbe outperforms bilingual models (English to Ewe and English to Fon) and gives state-of-the-art results on the JW300 benchmark for Fon established by \citet{nekoto2020participatory}. We hope this work will contribute to the massive adoption of Multilingual models inside the community. Our code is made accessible from \href{https://github.com/hgilles06/Englisg2Gbe}{Github}.
\end{abstract}

\section{Introduction}

    Language is one of the most important expressions of culture and implicitly of people's emancipation. The African continent is prosperous in more than 2,000 languages. However, most of them are low-resourced \citep{nekoto2020participatory}. Machine translation (MT) then emerged as a compelling technology to revive and strengthen African languages. Firstly, it could enable to break language barriers on the continent. Indeed, most African countries use languages inherited from colonization as official languages (such as English, French, Portuguese).
    Nevertheless, a non-negligible part of populations from the continent do not speak the official languages or at least not fluently. Official languages are learned at school, while sometimes most people do not have more than a primary school education level. This factor alone creates tremendous inequalities in African countries. First, it implicitly excludes a part of the society from participating in some crucial political debates and even some portion of the economy. Secondly, indigenous languages are often discarded from educational systems.
    
    Consequently, those who even successfully attend school can often speak indigenous languages while not being able to write or count in those languages. Cases are also noticeable where illiterate\footnote{Understand by \textit{illiterate} here, someone who did not attend at all conventional school or did not complete the primary school.} populations from the same country cannot chat together, often because they speak different indigenous languages. MT could allow us to at least decrease the information asymmetry created by this situation. Illiterates could better understand what is said in official documents if they could have an automatic translation. Furthermore, those who attended school and cannot write in their native languages could learn from such a tool. 
    
    Neural machine translation (NMT) is since several years the state-of-the-art methodology to deal with MT \citep{kalchbrenner2013recurrent, bahdanau2014neural}. NMT models were initially trained for specific language pairs. However, they have been extended to learn from several language pairs and translation directions within a single model: this is known as multilingual NMT \citep{dong2015multi, firat2016multi, ha2016toward, johnson2017google, aharoni2019massively}. Multilingual models are more efficient as they reduce the number of models to train and maintain while enabling transfer learning across languages\footnote{Mostly languages from the same family.}. This is even more appealing for low-resourced languages. \citep{johnson2017google, kocmi2018trivial, aharoni2019massively, arivazhagan2019massively}
    
    \begin{figure}[h]
      \centering
      \includegraphics[width=1\linewidth]{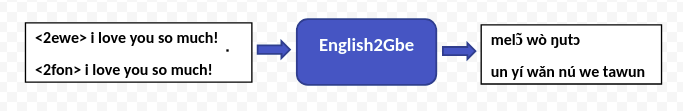}
        \caption{Translation from English to Ewe/Fon with a single model}
        \label{fig:translate}
    \end{figure}
    
    Despite the community's endeavors, lots of languages are still lagging behind the advancement of NMT. For instance, most African languages are excluded as they are often low and even very low-resourced. The work by \citet{nekoto2020participatory} has been a significant step toward accounting for African languages. The \href{https://www.masakhane.io/}{Masakhane} community provided an array of new datasets and more than 30 NMT benchmarks. 
    
\begin{figure}
      \centering
      \includegraphics[width=1\linewidth]{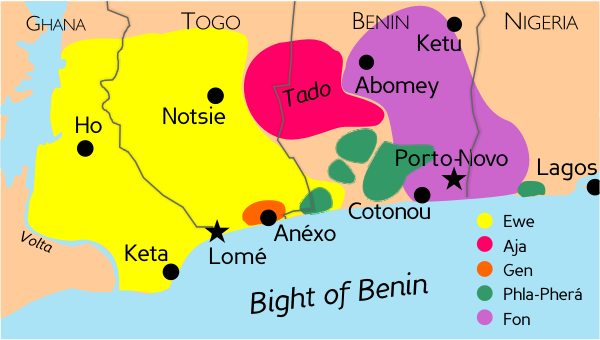}
      \caption{Geographical focus of Gbe languages}
      \label{fig:gbe}
      Source: Wikipedia
    \end{figure}

    A cluster of particularly low-resourced languages is the Gbe languages, which are widespread across Ghana, Togo, Benin, and Nigeria (see figure \ref{fig:gbe}).  The most known and most spoken ones are Ewe/Ewegbe and Fon/Fongbe. Approximately 4 and 2 millions people respectively speak them as native languages \citep{azunre2021nlp}  (see \href{https://en.wikipedia.org/wiki/Fon_language}{Wikipedia}). Ewe is the most spoken language in Togo, whereas Fon is the most spoken language in Benin. However, Ewe has less than 400 content pages on \href{https://ee.wikipedia.org/wiki/Special:Statistics}{Wikipedia}, and Fon is non-existent on the encyclopedia.  As both languages come from the same family \textbf{Gbe}, they share some phonetic segments, semantics, morphology and grammar rules \citep{capo2010comparative} (see \href{https://en.wikipedia.org/wiki/Gbe_languages}{Wikipedia}). Additionally, they are both tonal. 

    In this work, we want to showcase the advantage of using a multilingual NMT model compared to bilingual NMT models for these languages.
    Indeed,  a multilingual model could allow sharing of information between the languages resulting in better performance. A multilingual model also reduces the number of models to train and then to maintain. We trained two bilingual models, English to Ewe, English to Fon, and one multilingual model English to Fon/Ewe. We used the JW300 dataset \citep{agic2020jw300} and the benchmark proposed by \href{https://www.masakhane.io/}{Masakhane} \citep{nekoto2020participatory} to evaluate our models. 
    
    We used the Transformer model suggested by \citet{vaswani2017attention} and its implementation by \citet{kreutzer-etal-2019-joey} in order to facilitate reproducibility. To the best of our knowledge, we are the first to suggest an NMT model to translate from English to Ewe. Our bilingual NMT model from English to Fon gives a BLEU score about 11 points greater than the current state-of-the-art \citep{nekoto2020participatory}. Moreover, we showed that the multilingual model performs better than bilingual models, making it the new state-of-the-art. 
    
    The rest of the paper is structured in four sections. Section 2 presents some related work, while section 3 presents the data and methodology used. Finally, section 3 showcases our results, and the last section summarizes our main findings.

%\section{Low resource languages}

%Multilinguality

%Low ressource languages

\section{Related work}

\citet{capo2010comparative} is the groundbreaking work on finding common phonology features of Gbe languages. These common features are supposed to reflect the ones of the parent language. The most known Gbe languages are Ewe/Ewegbe, Fon/Fongbe, Gen/Gengbe, Aj\`a/Aj\`agbe and Gun/Gungbe. 
In this paper, we are only interested in the two most spoken ones: Ewe and Fon. 

To the best of our knowledge, the work by \citet{dossou2020ffr} is the first attempt to build a machine translation system accounting for Fon. The Fon-French Neural Machine Translation (FFR) project aimed to create a reliable Fon to French machine translation model. The second major work on a machine translation system accounting for Fon is the one of \citet{nekoto2020participatory}. This latter is generally the pioneering work on building large-scale machine translation models from English to African languages. An English to Fon model was part of the released models. 

To the best of our knowledge, there is no scientific work that built a machine translation system accounting for Ewe. However, there is an ongoing initiative from the \href{https://ghananlp.org/}{NLP Ghana organization} to account for Ghanaian languages, including Ewe \citep{azunre2021nlp}.
Moreover, there is no multilingual model accounting for Fon and Ewe to leverage their common features to the best of our knowledge. We are then the first to build such a model and showcase that having a single translation model from English to both languages is more beneficial. Not only because the number of models to train and to maintain is reduced, but also because the multilingual model performs better than single bilingual models.

\section{Data and Methodology}

We collected the English-Ewe and English-Fon language pairs datasets from the \href{https://opus.nlpl.eu/JW300.php}{JW300} corpus \citep{agic2020jw300}. The English-Fon dataset is made of 35,956 sentence pairs.  We did some filtering to assure that: 1) there are no duplicates; 2) there are no empty lines; 3) there are no test sentence pairs in the training set: we used the JW300 test set provided by Masakhane \citep{nekoto2020participatory}. After filtering, the size of the English-Fon dataset is 30,899 sentence pairs

The English-Ewe dataset is surprisingly huge and is made of 641,895 sentence pairs. After filtering, the size of the English-Ewe dataset is finally 580,350 sentence pairs. In order to well exploit our limited computing resources and to have a more balanced dataset for the multilingual model, we decided to limit the training set to 100,000 randomly selected sentence pairs. Indeed, our primary goal is to assess how much a multilingual model would benefit both languages (Ewe and Fon) compared to bilingual models. In table \ref{tab:en2ee}, we compared two (02) English to Ewe (English2Ewe) models: one trained on the complete training set\footnote{We used 1,000 observations as the validation set.} and another on the reduced set. The model trained on the complete set outperforms the one limited to 100,000 training observations in terms of BLEU score: 42.5 VS 35.7. However, the model with a limited training set achieves a good BLEU score. So, we decided to keep the limited training set for the rest of the paper. Details about the model architecture used for this comparison is given below.

Table \ref{tab:datasize} presents train, validation, and test set sizes for the three final models we trained for our analyses. 

Both bilingual models (English2Ewe and English2Fon) and the multilingual model (English2Gbe) are trained using the ``Base" Transformer architecture \citep{vaswani2017attention}. Here are the main and common parameters of models we trained: number of layers = 6; embedding size = 512; hidden size = 512; feed forward size = 2,048; number of attention heads = 8; maximum sequence length = 150. For bilingual models we used a batch size of 300. We trained
Byte-Pair Encoding (BPE) tokenizers \citep{sennrich2015neural} for each language and each model using the training data, except for Fon in the English2Fon model where we used additional scrapped data from the JW300 website. For Fon and Ewe, the vocabulary size used to train the BPE tokenizers is respectively set to 4,000 and 10,000 for Fon/Ewe and English. We chose a smaller vocabulary size for Fon/Ewe mostly because a single word in Gbe languages often has several meanings. Here is an example from \cite{gnanguenon2014analyse}: \textit{aw\u{a}}, which can have 4 completely different meanings in Fon. It could mean \textit{joy, big house, roof} or even \textit{carpenter tool} according to the context. This property, known as \textit{polysemous presentation} of Gbe languages, and generally reduces their vocabulary size compared to other languages such as English. 

%For information, there are: 1)  37,949 English (unique) tokens and 43,512 Ewe tokens in the training set for English2Ewe; 2) 15,322 English tokens and 7,075 Fon tokens in the training set for English2Fon. 

The batch size of the multilingual model is set to 400. We also learned BPE tokenizers from training sets. We still set the English vocabulary size to 10,000, while we now set it to 6,000 for Gbe languages (Ewe/Fon). We increased the vocabulary size for Gbe languages in the multilingual model as even if both languages share a substantial part of their phonology, they differ in vocabulary  \citep{capo2010comparative}. 

The exhaustive list of parameters can be found in our configuration files. Our code and dependent files are accessible from \href{https://github.com/hgilles06/Englisg2Gbe}{Github}.

Source sentences used in the multilingual model (in English) are tagged whether by $<2ewe>$ or $<2fon>$ at the beginning according to the target language (see figure \ref{fig:translate}). All the models are trained on 30 epochs to ease comparisons.

\begin{table}
    \centering
    \caption{Comparison of results from English2Ewe according to the training set size (using the Masakhane benchmark)}

    \begin{tabular}{ccccc}
        \hline
        \textbf{Model} & \textbf{Training set size} & \textbf{BLEU}   & \textbf{CHRF}  & \textbf{TER} \\
        \hline
        English2Ewe & 579,350 & 42.5 &  61.2    &  44.5 \\
        English2Ewe & 100,000 & 35.7 &  54.9    &  51.1 \\
        \hline
        \end{tabular}
        
    \label{tab:en2ee}
    
     \begin{tablenotes}
        \centering
        \small
        \item Note: English2Ewe: English to Ewe. We used 1,000 sentence pairs for the validation set.
    \end{tablenotes}
\end{table}

\begin{table}

\centering
\caption{Dataset sizes}
\begin{tabular}{llll}
    \hline
     \textbf{Model}   & \textbf{Train}   & \textbf{Validation} & \textbf{Test}  \\
    \hline
    English2Ewe & 100,000 & 1,000      & 2,720  \\
    English2Fon & 29,899  & 1,000      & 2,718  \\
    English2Gbe & 129,889 & 2,000      & 5,438 \\
    \hline 
    \end{tabular}
    \label{tab:datasize}
    
    \begin{tablenotes}
        \centering
        \small
        \item Note: English2Ewe: English to Ewe; English2Fon: English to Fon; English2Gbe: English to Ewe/Fon. BLEU, CHRF, and TER scores are computed using the Sacrebleu Python package.     
    \end{tablenotes}
    
\end{table}

\section{Results} \label{sec:results}

Table \ref{tab:results} presents evaluation results for bilingual models (English2Ewe and English2Fon) and for the multilingual model (English2Gbe) using the JW300 test set provided by Masakhane \citep{nekoto2020participatory}. We computed the BLEU and the CHRF using the Sacrebleu Python package to make results comparable. We only compare our results for Fon to the ones from \cite{nekoto2020participatory}  as they do not report any results on Ewe.  For English2Fon, we reported a BLEU score of 42, whereas the state-of-the-art was 31.07. Results are even better with English2Gbe: we reported a BLEU score of 44.8 on the Fon test set. 

For English2Ewe, the BLEU score on the test set is 35.7, which is an acceptable score for 30 epochs. As we have more data on Ewe, increasing the number of epochs should allow a better score. Nevertheless, for the sake of this work, we limit the training to 30 epochs as our primary goal is to compare bilingual models to the multilingual one in the same configuration context.  English2Gbe tested on Ewe gives slightly better BLEU, CHRF, and TER scores compared to English2Ewe. 

Roughly speaking, the multilingual model improves translation quality for Fon while preserving translation quality for Ewe. This is coherent with results from \citet{aharoni2019massively}.

\begin{table}

    \centering
    \caption{Evaluation results for translations from English to Gbe languages (Fon/Ewe) using the Masakhane benchmark}
    
    \begin{tabular}{ccccc}
        \hline
        \textbf{Model} & \textbf{Target} & \textbf{BLEU}   & \textbf{CHRF}  & \textbf{TER} \\
        \hline
        English2Ewe & Ewe & 35.7 &  54.9    &  51.1 \\
        English2Fon & Fon  & 42 & 54.1   &  48.4 \\
        \multirow{3}{*}{English2Gbe} & Gbe (Ewe/Fon) & 41.2  & 56.6  &  46.5 \\
                                     & Ewe &  36.1  &  55.8 & 49.9 \\
                                     & Fon &  44.8  &  56.9 & 43.9 \\
    
        \hline
        \end{tabular}
        
        \label{tab:results}
        
        \begin{tablenotes}
            \centering
            \small
            \item Note: English2Ewe: English to Ewe; English2Fon: English to Fon; English2Gbe: English to Ewe/Fon.  
        \end{tablenotes}
    
\end{table}

\section{Conclusion}

The main goal of this paper was to showcase the benefit of using multilingual neural machine translation (NMT) models instead of bilingual ones, especially for low-resourced languages. This paper introduced English2Gbe, a multilingual NMT model capable of translating from English to Ewe or Fon. English2Gbe provides state-of-the-art results on the JW300 benchmark for Fon proposed by Masakhane \citep{nekoto2020participatory}. 
To the best of our knowledge, we are the first to provide an official and open source translation system for ``English to Ewe" translation. Furthermore, we showed through automatic scores (BLEU, CHRF, and TER) that the multilingual model outperformed bilingual ones.

Our code is accessible from \href{https://github.com/hgilles06/Englisg2Gbe}{Github} to allow reproducibility and can serve as a baseline for future works from the community. Training and test sets, but also configuration files, are as well made available.

%\section*{Acknowledgements}
%We are grateful Sebastian  to Julia Kreutzer for her constructive comments.

% Entries for the entire Anthology, followed by custom entries
\bibliography{custom}
\bibliographystyle{chicago}

\end{document}